\documentclass[pmlr]{jmlr}

\usepackage{tcolorbox}
\tcbuselibrary{breakable}
\usepackage{xcolor}
\usepackage{appendix}
\usepackage{booktabs}
\RequirePackage{graphicx}
\usepackage{tabularx}
\usepackage[labelfont=bf]{caption} 

\usepackage{subcaption} 
 \usepackage{booktabs}
\setcitestyle{numbers}

\usepackage{longtable}
\usepackage{makecell}

\makeatletter
\def\set@curr@file#1{\def\@curr@file{#1}} 
\makeatother
\usepackage[load-configurations=version-1]{siunitx} 

\title[Ten Headache Specialists versus Artificial Intelligence for Clinical Literature Summarization]{Ten Headache Specialists versus Artificial Intelligence for Clinical Literature Summarization: A Critical Evaluation and Comparison}

\author{
\Name{Alejandro Lozano}*$^{1}$
\Name{Keiko Ihara*$^{2}$}
\Name{Ping-Hao Yang$^{3}$}
\Name{Carrie E. Robertson$^{4}$} \\
\Name{Jennifer Stern$^{5}$}
\Name{Allan Purdy$^{6}$}
\Name{Hsiangkuo Yuan$^{7}$}
\Name{Pengfei Zhang$^{8}$}
\Name{Yulia Orlova$^{9}$}\\
\Name{Olga Fermo$^{10}$}
\Name{Jennifer Hranilovich$^{11}$}
\Name{Fred Cohen$^{12}$}
\Name{Todd J. Schwedt$^{2}$}\\
\Name{Jenelle A. Jindal$^{1}$}
\Name{Serena Yeung-Levy$^{1}$}
\Name{Chia-Chun Chiang$^{\dagger}$$^{2}$} \\ \\
\addr $^{1}$ Stanford University, Palo Alto, CA, USA\\
\addr $^{2}$ Department of Neurology, Mayo Clinic, Rochester, MN, USA\\
\addr $^{3}$ Department of Neurology, Dalhousie University, Halifax, Canada\\
\addr $^{4}$ Jefferson Headache Center, Department of Neurology, Thomas Jefferson University, PA, USA\\
\addr $^{5}$ Beth Israel Deaconess Medical Center, Boston, MA, USA\\
\addr $^{6}$ Department of Neurology, University of Florida, Gainesville, FL, USA\\
\addr $^{7}$ University of Colorado School of Medicine, Department of Pediatrics, Division of Child Neurology, Aurora, CO, USA\\
\addr $^{8}$ Department of Medicine, Mount Sinai Hospital, Icahn School of Medicine at Mount Sinai, New York, NY, USA\\
\addr $^{9}$ Department of Neurology, Mayo Clinic, Scottsdale, AZ, USA\\
\addr $^{10}$ Harvard Medical School, Boston, MA, USA\\
\addr $^{11}$ Department of Neurology, Mount Sinai Hospital, Icahn School of Medicine at Mount Sinai, New York, NY, USA\\ \\
\addr $^{*}$ Denotes equal contribution. \\ 
\addr $^{\dagger}$ Denotes corresponding authors: Chiang.Chia-Chun@mayo.edu
}

\begin{document}

\maketitle

\begin{abstract}

\noindent \textbf{Background:}Summarizing the latest medical literature to guide clinical decision-making is essential for evidence-based medicine and high-quality patient care. Yet clinicians face increasing challenges due to limited time with patients and a staggering growing volume of published articles. Although retrieval-augmented large language models (LLMs) have shown promise in clinical summarization, human evaluations of their effectiveness in synthesizing broader scientific literature and direct comparison to expert-written synthesis remain scarce.

\noindent \textbf{Methods:}We constructed a RAG-based agentic AI framework using three state-of-the-art LLMs, Sonnet, GPT-4o, and Llama 3.1. A headache specialist created 13 questions: three for prompt optimization and ten for evaluation. Ten headache specialists across the United States and Canada each wrote a summary for one question, yielding four summaries per question (expert, Sonnet, GPT-4o, and Llama). The experts, blinded to authorship, critically evaluated the summaries, excluding the topic for which they wrote a summary, based on correctness, completeness, conciseness and clinical utility, scoring from 1-10 with standardized rubrics provided. They ranked the summaries by preference and indicated which they believed was written by an expert vs LLM.

\noindent \textbf{Results:}Two hundred critical assessments of the summaries from ten experts were analyzed. Overall, human experts’ written summaries scored the highest in all evaluation metrics and were the most preferred responses, followed by Sonnet, GPT-4o, and Llama 3.1. Paired comparisons showed headache specialists outperformed GPT-4o and Llama on most metrics, with no significant difference compared to Sonnet. Headache specialists correctly distinguished expert-written from AI-written summaries 64\% (32/50) of the time. Beyond the metrics used, experts valued the quality of references, inclusion of key clinical details such as medication dosage, synthesis across sources, and incorporation of clinical nuance and experience.

\noindent \textbf{Conclusion:} Our study, comparing LLM and expert-written literature summaries evaluated by headache specialists, showed that expert-written summaries were preferred, though it was sometimes challenging for experts to distinguish between human vs AI-generated ones. We also identified key expert-valued features, beyond standard evaluation metrics, that can guide future refinement of human and AI literature summarization pipelines.

\end{abstract}

\section{Introduction}
\label{sec:intro}

Evidence-based medicine, defined as the conscientious, explicit, and judicious use of current best evidence in making decisions about the care of individual patients \cite{sackett1996evidence} has long been regarded as the gold standard of clinical practice. However, summarizing and integrating the latest literature for each patient during increasingly time-constrained visits has become a significant challenge \cite{tai2007time} as scientific literature grows exponentially each year. Resources such as UpToDate are widely used to address this gap, relying on expert clinicians to synthesize and summarize the medical literature for use at the point of care. Yet, novel evidence from early-stage therapies, emerging clinical trials, and case studies often develops more rapidly than can be incorporated into expert-curated repositories. Furthermore, a broad review of a topic might not provide sufficient details suitable for a specific clinical scenario, highlighting opportunities for improvement.

Large language models (LLMs) have shown promise across a wide range of clinical tasks, including diagnostic support \cite{goh2024large}, drafting discharge summaries \cite{zaretsky2024generative}, and retrieving information from electronic health records \cite{fleming2023medalign}. Building on this success, interest has grown in LLM-based systems capable of retrieving, evaluating, and synthesizing scientific literature with expert-level proficiency on demand, a trend that has steadily matured in recent years. We have previously built an agentic AI framework based on chains of retrieval-augmented LLMs, clinfo.ai, for summarizing medical literature from resources such as PubMed. The framework was made open source \cite{clinfo}. Additionally, in May 2025, the U.S. Food and Drug Administration launched its first LLM-assisted scientific review pilot. Similarly, as of July 2025, OpenEvidence, an AI-powered platform that aggregates and synthesizes peer-reviewed medical literature, reported onboarding 65,000 new verified U.S. clinicians per month.

Despite the widespread use of LLMs to summarize clinical evidence, several important questions remain unanswered. How do LLM-generated summaries compare with those written by clinical experts on specialized topics, and what characteristics distinguish expert-authored summaries from those produced by LLMs? What do clinical experts emphasize when writing and evaluating literature summaries, and what key attributes, beyond commonly used LLM evaluation metrics, define a high-quality summary from their perspective? Addressing these questions is important not only for assessing the current capabilities of LLMs, but also for the development of future pipelines using LLMs to summarize and synthesize literature. 
From our previous experiences evaluating LLM outputs for medical tasks, we found that there was substantial variation and low agreement across participating clinicians (Fleming et al. 2023). This may be related to the substantial variability across different medical specialties. To pursue high quality, expert-level assessments, conducting such evaluations within a specific area of medicine is crucial. We focused on Headache Medicine for this study, designing and optimizing the clinfo.ai framework with content experts to enhance the quality of the generated summaries. We conducted a multidimensional clinical evaluation comparing ten headache specialist versus AI-generated summaries from different LLM systems across seven axes— correctness, completeness, conciseness, clinical usefulness, preference, identification of human-written summaries, and free text comments. The evaluations covered 15 subcategories within these dimensions. Our assessment is extensive and detailed, totaling over 120 annotation hours by a panel of 10 academic headache specialists with an average of 21.9 years of medical experience across the United States and Canada to critically evaluate all summaries. In total, each headache specialist reviewed 5 questions (20 summaries) and 180 annotated items (36 items per question), resulting in 200 critically evaluated summaries generated by LLMs or written by experts, and a total of 1,800 expert-annotated concepts. To complement our multidimensional evaluation, we additionally performed an in-depth free-text assessment to capture nuanced strengths and weaknesses of the generated summaries (missed by commonly used metrics), emphasizing aspects that clinical experts value in retrieval-augmented summarization.
Our two-part evaluation shows that standard quantitative metrics alone were not sufficient to capture expert preferences, revealing limitations in prior assessments of LLM outputs. Even when clinicians cannot reliably distinguish human- from AI-generated summaries, they still prefer human-written ones, by a margin nearly twice that of the best-performing model (Sonnet). Clinicians identified literature misinterpretation, omission of key concepts, and missing critical references as major bottlenecks of current RAG-enabled LLMs.

\section{Related Work}

\subsection{Large Language Models (LLMs)}

Large language models (LLMs) are deep learning systems trained on Internet-scale datasets that exhibit strong natural language processing capabilities \citep{brown2020language, kaplan2020scaling}. In medicine, LLMs have been applied to conduct diagnostic conversations with patients \citep{tu2024towards,mirza2024using}, summarize and generate clinical notes, support clinical trial matching \cite{wornow2025zero}, analyze and synthesis unstructured electronic health record data, \cite{fleming2023medalign} and provide decision support for complex cases \cite{wu2025medcasereasoning}. 

Remarkably, LLMs can perform all these tasks without task-specific training \citep{brown2020language}—a phenomenon known as zero-shot learning \citep{xian2018zero}. This capability stands in stark contrast to traditional machine learning models, which depend on labeled datasets, and to rule-based systems, which require domain experts to manually encode rules and logic.

\subsection{Retrieval Augmented Langue Models}

LLMs often produce factually incorrect or unverifiable statements, a phenomenon commonly referred to as hallucination \citep{huang2025survey}. This limitation poses a particular challenge in medicine, where factual grounding and source traceability are critical. This gap has led to the growing adoption of retrieval-augmented generation (RAG) approaches, in which relevant external documents (e.g. scientific literature) are dynamically retrieved and incorporated into the model’s context before generation. This allows LLMs to produce answers supported by explicit and traceable evidence rather than relying solely on parametric memory.

To this end, several studies have demonstrated the effectiveness of retrieval-based approaches for clinical question answering and decision support. For instance, Almanac \cite{zakka2024almanac} and Clinfo.ai \citep{lozano2023clinfo} (our previously developed framework) are open-source systems that retrieve biomedical literature from PubMed to answer clinical questions, providing both natural language summaries and traceable citations. Similarly, domain-specific retrieval-augmented models have been developed for specialized areas of medicine: Guide-Bot integrates retrieval from medical guideline repositories to enhance factual accuracy in oncology \citep{lee2024development}, while DALK \cite{li2024dalk} is tailored to answer questions related to Alzheimer’s disease using scientific literature.


 \subsection{Evaluation of Retrieval Augmented Language Models for literature summaries}

Given the growing development of multiple frameworks to summarize medical evidence using  RAG LLMs, there is a growing need for systematic evaluations. Medeviedence \cite{polzak2025can} introduces a closed VQA benchmark designed to assess retrieval performance on systematic reviews. Their findings show that current models underperform relative to human experts and often exhibit unjustified confidence and insufficient skepticism. However, benchmark performance alone does not provide a complete picture; expert evaluation remains essential. To this end, Tang et al. (2023) \cite{tang2023evaluating} examined the ability of two LLMs to summarize medical evidence across six clinical domains using 53 systematic reviews, including 10 in neurology. Clinical experts rated each generated summary for coherence, factual accuracy, comprehensiveness, and potential for harm. Subsequent work expanded this line of evaluation to broader clinical tasks: Luo et al. (2024) \cite{luo2024development} studied a RAG system for ophthalmology using a panel of six medical experts, while Ke et al. (2025) \cite{ke2025retrieval} assessed RAG systems for determining surgical fitness and providing preoperative guidance, focusing on accuracy, consistency, and patient safety. Across these studies, retrieval augmentation consistently reduced hallucinations, improved factual correctness, and increased clinician trust relative to LLM-only baselines.\textbf{Yet, no prior work directly compares RAG-enabled LLM outputs to those of expert practicing clinicians.} Consequently, we lack a clear understanding of how these systems perform and contrast with respect to  experts. Since outperforming standard LLMs is insufficient for real-world deployment, addressing this gap is essential for building reliable, clinically useful systems.



\section{Methods}

\subsection{Models}
\label{sec:llms}

For our assessment, we include the proprietary models GPT-4o (OpenAI) \citep{gpt4} and Sonnet (Anthropic), along with the widely used open-source model Llama-3.1 \citep{peng2023yarn}. The open-source models are executed on a local PHI-compliant server equipped with eight NVIDIA H200 GPUs using vLLM \citep{vllm}. This selection is not comprehensive but rather a focused subset designed to represent both closed and open models given the time constrains of expert clinical evaluation.

\subsection{Retrieval Pipeline}

We employed the starter retrieval pipeline initially proposed by \cite{lozano2023clinfo}, incorporating modifications that were rigorously validated in  \cite{polzak2025can}. System optimization was conducted in collaboration with a headache specialist (CC) with extensive clinical and scientific writing experience. Prompts were iteratively refined using a held-out dataset comprising three representative questions spanning epidemiology, diagnosis, and treatment considerations across different headache disorders.
Prompts were engineered in multiple steps and aimed to optimize article selection by prioritizing higher levels of evidence, providing the different levels of evidence for different study types, and instructed the model to include original research articles rather than citing narrative reviews. The prompt also included instructions to include clinically actionable details, such as medication names, dosages, and routes of administration.
Our agentic, chain-of-LLMs pipeline integrated multiple LLMs with a scientific article search index (PubMed). It comprised five sequential steps:

\begin{enumerate}
    \item \textbf{Query Generation}: Transforms the user's question into a carefully constructed search query using MeSH terms.
    \item \textbf{Information Retrieval}: Fetches abstracts from PubMed using a search index.
    \item \textbf{Relevance Classification}: Employs an LLM to determine whether each retrieved article is relevant.
    \item \textbf{Summarization}: Condenses each relevant abstract, highlighting information most pertinent to the user's question.
    \item \textbf{Synthesis}: Integrates the summarized abstracts into a structured list and generates a concise, evidence-based summary.
\end{enumerate}

\subsection{Dataset collection}
Ten headache specialists from North America (the U.S. and Canada) were invited to participate in a two-phase study. In the first phase, clinicians were asked to formulate a clinical question and provide a 200–300-word response in a review-article format, including citations (Supplemental File). They were instructed to be concise and written for practicing clinicians and medical trainees to guide clinical practice and medical education. The participating experts were all headache specialists from academic headache centers with an average of 21.9 years of medical experience post medical school graduation, and were all actively engaged in research. Each had prolific experience writing review articles, with a minimum of three published narrative or systematic reviews.

\subsection{Evaluation by Headache Specialists}

In the second phase, an online questionnaire (Supplemental File) was distributed to the ten headache specialists who participated in the first phase of this study. Each headache specialist reviewed all summaries for five questions that they were not assigned to in Phase 1. The questionnaire included a rubric-based evaluation across the following categories and subcategories:

\begin{enumerate}
    \item \textbf{Correctness} (hallucinated or fabricated content, misinterpretation, fabricated citations)
    \item \textbf{Completeness} (missing important concepts, incomplete statements, absence of key references)
    \item \textbf{Conciseness} (overly long summaries, inclusion of unnecessary information, redundant content)
\end{enumerate}

We separated each comment from reviewers into phrases so that each phrase can represent one of the subcategories. Each category was scored on a scale from 1 (worst) to 10 (best). When flaws were identified, one point was deducted from 10 for each flaw within the relevant subcategory. Additionally, participants were asked to rate the overall usefulness of each summary on a scale from 1 (not useful) to 10 (extremely useful), to identify the most preferred summary among the four provided, and to indicate which summary they believed was written by a human expert. Of note, there were four summaries for each question, one written by an expert, three by different LLMs on the developed framework (GPT, Llama and Sonnet). Free-text responses were also collected to capture the rationale for their selections and any additional comments that did not fit within the structured rubric. These qualitative responses were reviewed by two headache experts (CC, KI) and analyzed thematically. All free text comments were read and categorized. The phrases/sentences of these comments were classified into the 12 topics relating to the reasons of why the reviewers identified human expert written answers vs AI generated summaries, and 14 topics among optional free comments, respectively. The number of responses that included each topic was summarized. When the reviewers explicitly stated that they had comments that did not fall into the rubric categories (correctness, completeness, and conciseness) but mentioned significant elements, those were added to the pool of answers for [Optional] questions.

\section{Results}

\begin{figure}
    \centering
    \includegraphics[
        width=\linewidth,
        trim=0 0 0 1cm,
        clip
    ]{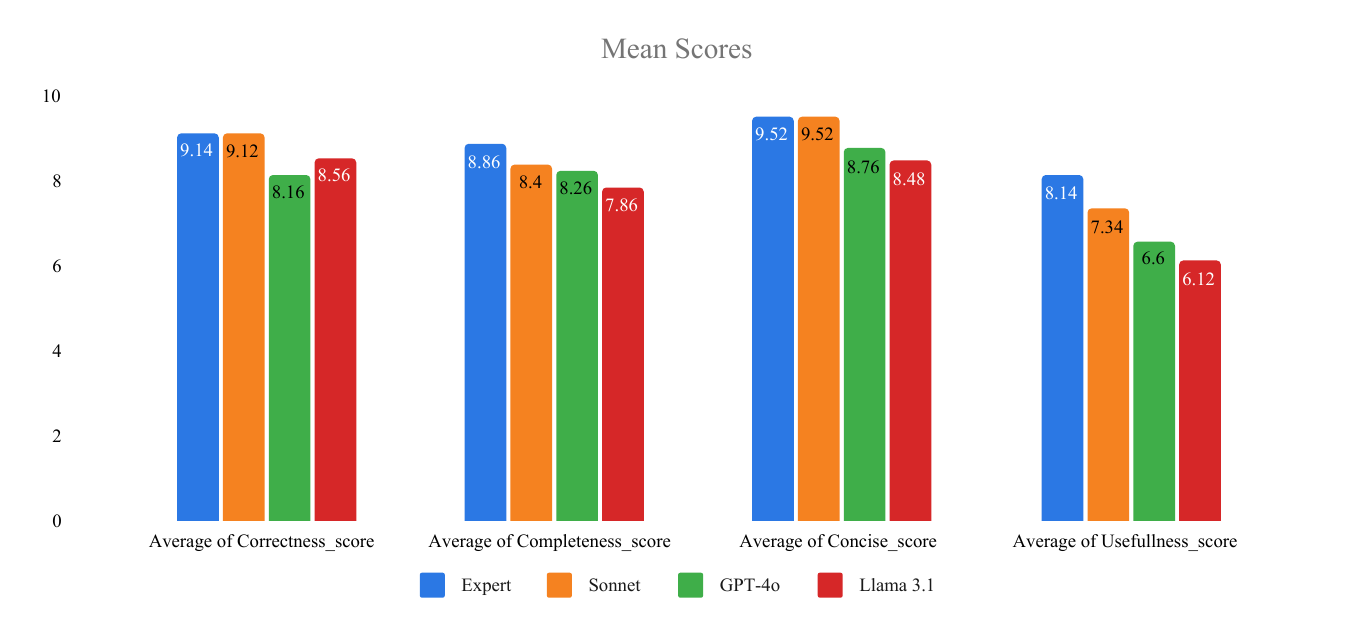}
    
    \caption{Average expert-assigned scores (0–10) across correctness, completeness, conciseness, and usefulness for retrieval-augmented answers/summaries to all ten questions. The maximum score of 10 indicates that no points were deducted in any category for summaries generated by the LLM or human experts.}
    \label{fig:1}
\end{figure}

We received 50 evaluations, 200 critically evaluated summaries, and 1800 annotated concepts from ten headache specialists, spanning 120+ hours of detailed review. Each response included 5 numerical values (correctness, completeness, conciseness, usefulness and preference) and 5 text responses (reasons for scoring correctness, completeness, conciseness, and human-written summary identification and optional comment). There were no missing data for the numerical entries, although a total of 13/150 text responses for correctness, completeness, and conciseness were empty. We received 50 comments relating to their choice of identifying human authors among AI generated summaries, and 23 optional free-text feedback.

\subsection{Correctness, Completeness, Conciseness, Usefulness, Preference}
Overall, comparison by analysis of variance among the four types of summaries (clinical expert, Sonnet, GPT-4o, and Llama 3.1) showed significant differences in all four 10-point scales (correctness, p=0.009; completeness, p=0.015; conciseness, p<0.001; usefulness, p<0.001). Expert-written summaries received the highest scores across correctness, completeness, and usefulness with averages of 9.14, 8.86, and 8.14, respectively (Figure 1), and were the most preferred by evaluators (26/50), averaging 3.1 in ranking (compared to 2.68 Sonnet, the second-best model). Regarding conciseness, expert-written and Sonnet-generated summaries were tied for the highest scores at 9.52.

\begin{figure}
    \centering
    \includegraphics[
        width=\linewidth,
        trim=0 0 0 3cm,
        clip
    ]{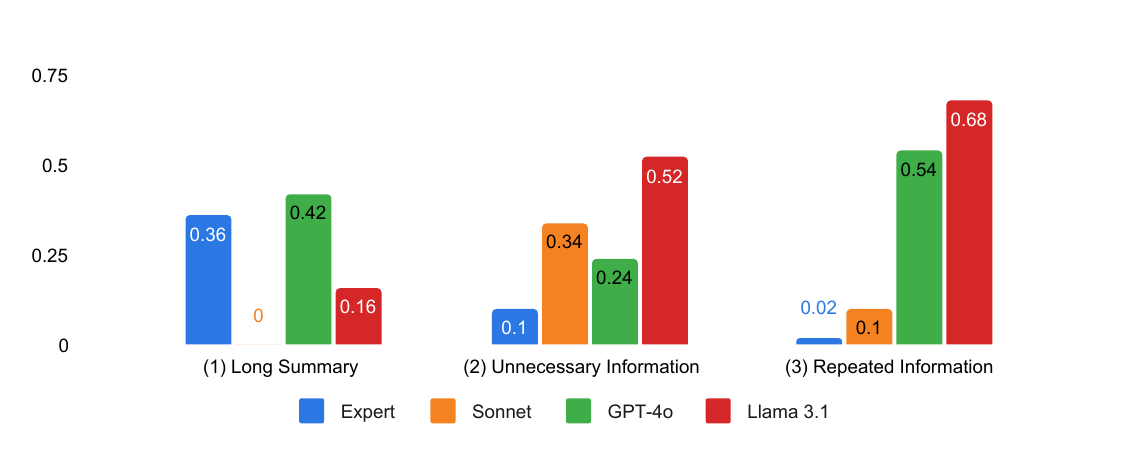}
    \caption{Human evaluation of summary conciseness. Mean point deductions assigned by headache specialists for three conciseness-related criteria: (1) overly long summaries, (2) inclusion of unnecessary information, and (3) repeated information. Higher scores indicate larger penalties and therefore poorer conciseness. Expert-written summaries received the fewest penalties overall, whereas Llama 3.1 was most frequently penalized for unnecessary and repeated information.}
    \label{fig:conciseness}
\end{figure}

Comparison between expert summaries and each LLM output showed that for all four categories, expert-written summaries and Sonnet-generated summaries had no significant difference (correctness 9.14 vs 9.12, p=0.947; completeness 8.86 vs 8.40; p=0.130; conciseness 9.52 vs 9.52; p=1.0; usefulness 8.14 vs 7.34, p=0.057). Expert-written summaries scored significantly higher than GPT-4o in all four categories (correctness 9.14 vs 8.16, p=0.013; completeness 8.86 vs 8.26, p=0.025; conciseness 9.52 vs 8.76, p=0.008; usefulness 8.14 vs 6.60, p<0.01) and scored significantly higher than Llama 3.1 in completeness (8.86 vs 7.86, p<0.001), conciseness (9.52 vs 8.48, p<0.001), and usefulness (8.14 vs 6.12, p<0.00). However, for correctness, there was no statistically significant difference between expert and Llama 3.1 (9.14 vs 8.56, p=0.059).

Examining the nine rubric subcategories, expert-written summaries had the lowest point deductions (i.e., best performance) across all five subcategories (average scores): correctness—misinterpretation (0.18); completeness—lack of important concepts (0.3) and lack of key references (0.22); and conciseness—unnecessary information (0.1) and repeated information (0.02). Sonnet-generated summaries had the lowest deductions in three subcategories (average scores): correctness—hallucinated/fabricated content (0.3) and fabricated citations (0); conciseness—long summaries (0). GPT-4o had the lowest deductions for completeness-incomplete statement with an average of 0.42. Llama 3.1 did not demonstrate the best performance in any of the four major categories or the subcategories.

\begin{figure}
    \centering
     \includegraphics[
        width=\linewidth,
        trim=0 0 0 70 cm,
        clip
    ]{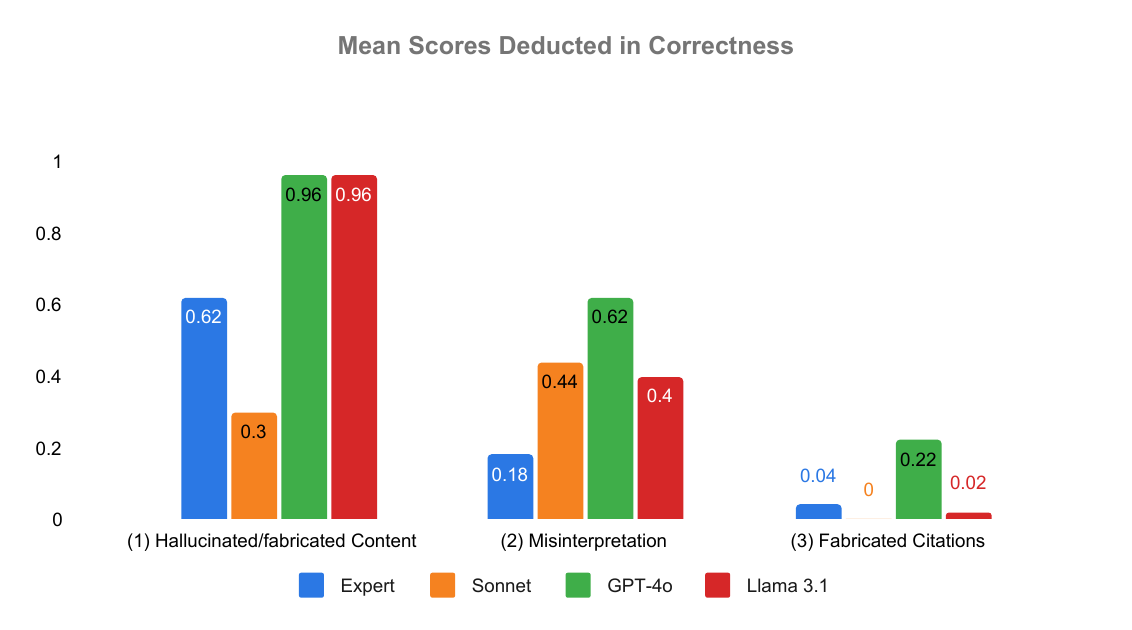}
    \caption{Human evaluation of summary correctness. Mean deduction scores assigned by headache specialists for three correctness-related criteria: (1) hallucinated or fabricated content, (2) misinterpretation of the cited literature, and (3) fabricated citations. Higher scores indicate larger penalties and therefore lower factual correctness, whereas lower scores indicate greater faithfulness to the source literature. GPT-4o and Llama 3.1 received the largest penalties for hallucinated or fabricated content, while fabricated citations were relatively uncommon across all methods.}
    \label{fig:placeholder}
\end{figure}

\subsection{Identification of Expert Written Summaries and Additional Feedback}

Out of 50 responses (5 per question), 32 (64\%) correctly identified the expert-written summaries from LLM-generated summaries. Among the 18 incorrect identifications, 9, 7, and 2 responses incorrectly identified Sonnet-, GPT-4o-, and Llama 3.1- generated summaries as human-written summaries, respectively. The most frequently attributed features to expert-written summaries were focus of information (n=14), completeness (n=12), reference quality (n=12), conciseness (n=10), expert opinion and judgment (n=8), and flow of information (n=8). Additional features mentioned included AI mistakes such as incorrectly claiming the summary was a systematic review (n=4), human errors including grammar errors or typos (n=4), readability and writing style (n=4), synthesis (n=3), up-to-date information and full-text content (n=3), and correctness (n=3).
Notably, four features outside of rubric were mentioned only by reviewers who correctly identified the expert-written summaries: correctness, synthesization, AI mistakes, and up-to-date information/full-text content. Synthesization of available information was described as “rather than quoting the available data, this author seemed to be trying to help readers interpret the data as a clinician.” The experts also noted that the types of errors in the summary were different between AI and experts. While some noted the "lack of mistakes" and "absence of errors and omission," others emphasized critical errors that human writers never do, including incorrectly referring "to itself as a review or a systematic review." Other features were reported by both participants who correctly identified the summaries as expert-written and those who did not. Several factors were mentioned more often in correct responses, including completeness (n=8 vs n=4), flow (n=5 vs n=3), focus (n=10 vs n=4), expert opinion/judgment (n=7 vs n=1), human errors (n=3 vs n=1), reference quality (n=8 vs n=4), and readability/writing style (n=3 vs n=1). However, conciseness was mentioned by 5 correct and 5 incorrect responses.
We additionally received 23 optional comments outside the rubric evaluation form and human identification. Features uniquely highlighted in these responses included the number of references per argument (n=3), use of incorrect sources or clinically irrelevant information (n=2), and authority of the reference source (n=2). Features overlapping with the rubric included focus of information (n=6), reference quality (n=6), conciseness (n=5), availability of English references (n=5), flow of information (n=4), synthesization (n=3), human writing style (n=2), up-to-date information (n=1), and expert opinion/judgment (n=1).
Importantly, several participants noted that the most helpful summary was not always the one they identified as expert-written. For example, in response to Q8, one participant wrote: “Although I chose summary B [expert-written] as written by a person, I thought that summary D [GPT-4o generated] was perhaps more useful as it gave dosing ranges.”
Discussion
We optimized an agentic chain-of-LLMs pipeline to efficiently summarize published medical literature in response to user queries. We then rigorously evaluated summaries generated by three LLMs (Sonnet-, GPT-4o-, and Llama 3.1) using this pipeline against those written by ten experienced headache specialists. Evaluated through rubric metrics, human-written summaries outperformed AI by a small margin, though not consistently across all subcategories. Headache specialists correctly identified 64\% of human-written summaries, and 52\% rated them as their top preference. Overall, Sonnet performed comparatively to human written summaries and was ranked the second most favored summary, while GPT-4o and Llama scored significantly lower for three rubric categories.

\begin{figure}
    \centering
     \includegraphics[
        width=\linewidth,
        trim=0 0 0 70 cm,
        clip
    ]{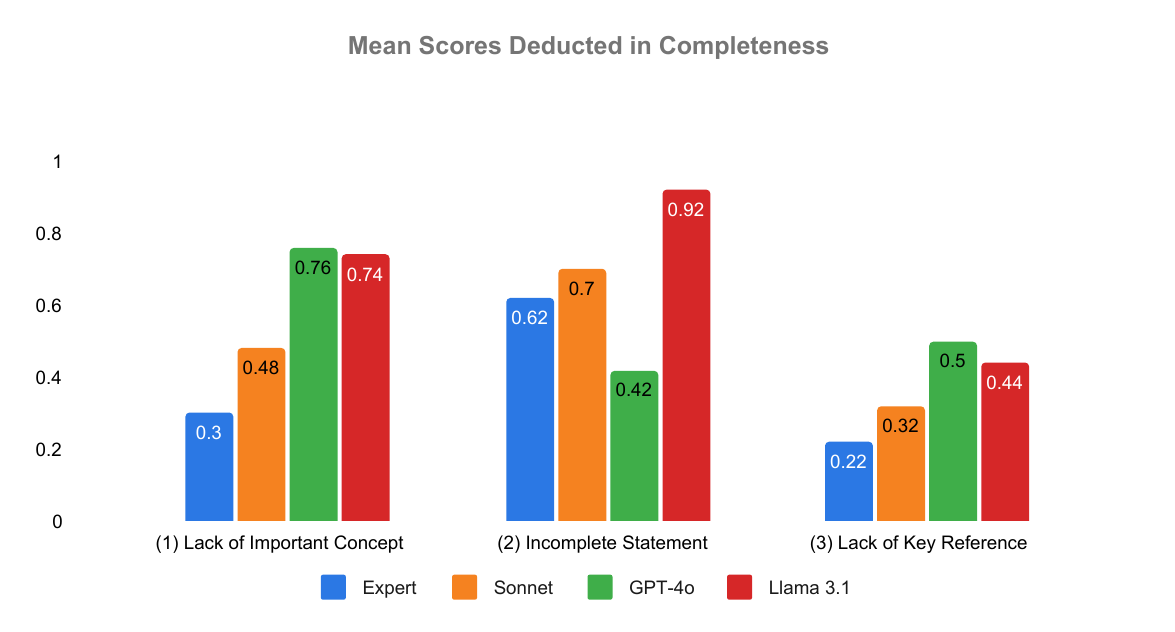}
    \caption{Human evaluation of summary completeness. Mean deduction scores assigned by headache specialists for three completeness-related criteria: (1) omission of important concepts, (2) incomplete statements, and (3) omission of key references. Higher scores indicate larger penalties and therefore lower completeness, whereas lower scores indicate more comprehensive summaries. Llama 3.1 and GPT-4o received the largest penalties for missing important concepts, while Llama 3.1 was most frequently penalized for incomplete statements.}
    \label{fig:placeholder}
\end{figure}

\section{Discussion}

We analyzed rubric and free text responses by 10 headache experts on LLM generated summaries and compared the feedback among human-written summaries, Sonnet-, GPT-4o-, and Llama 3.1- generated summaries. Overall, using a rubric,  human-written summaries outperformed AI by a small margin , though not consistently across all subcategories. Sixty-four percent of responses correctly identified the human-written summaries, and 52\% rated them as their top preference. Different LLM models showed different strengths and weaknesses. Overall, Sonnet performed comparatively to human written summaries and was ranked the second most favored summary, while GPT-4o and Llama scored significantly lower for three rubric categories. 

\subsection{Beyond metrics, free-text evaluation}

While AI generated summaries are close to expert-generated summaries when using only systematical evaluation metrics (i.e., correctness, completeness, conciseness, usefulness), some experts mentioned the weakness of metrics (e.g., "using the metrics is very time consuming and not always helpful," "the metrics only get you so far"). Thematic analyses of free-text responses revealed further detailed information on what attributes are valued in summaries: Some comments mentioned structural elements such as flow, focus, synthesization, readability and writing style; while others also listed the quality of contents, including expert opinion/judgment, reference quality, up to date information/full-text content, availability of English references, number of references, and credit/authority of reference source, for example, the inclusion of milestone phase-3 randomized controlled trials that established the efficacy of a treatment for headache disorders, or the International Classification of Headache Disorders diagnostic criteria, which is the gold standard reference for the diagnosis of headache disorders. We will elaborate on each of these preferences below.

\subsubsection{Structural elements: Flow, Focus, and Synthesization}
Experts have preferences for specific kinds of summary structure: It was important to have flow that is "similar to how a clinician thinks about treatment options," which can be outlined as "introducing the topic, then giving specifics, and concluding with a summary of how to initially evaluate such patients." Starting from very basic contents such as quotes from "the ICHD criteria" and grouping treatment types "by medical then surgical management" were considered ideal. Furthermore, experts value specific clinically relevant information. They appreciated "useful dosing information," "key clinical concepts," "controversies relevant to clinical practice," and "contraindications". When the information was "very superficial," "provided rare details and focused too much on diagnosis," and skewed "towards specific diagnosis," summaries were not seen as useful. An emphasis on the right perspective (e.g., "I was looking for an APPROACH and not therapy," "one of the samples was oddly surgery-heavy") and providing "balanced" information was important. Summaries "written by someone who could anticipate some of the questions that a reader might have in clinic" were considered valuable.
Synthesis, described as using "inference to summarize some of the citations" rather than "quoting the available data" or "’cut and pasted’ sentences from the references" appeared to be human-like and was valued. For example, when "citation says mesencephalon, summary says midbrain", it was seen as evidence for human writing, while "just listing stuff rather than making recommendations" was seen as LLM-like.

\subsubsection{reference quality and authority}

Overall, high-quality references were described as citations that include “a major meta-analysis, showing it to be evidence-based, major guidelines for specialists and non-specialists.” Additionally, it appeared to be important to have outstanding credit/authority of reference source represented as “the most important individuals in the headache world involved in the referenced material, and from great journals.” Besides quality, the number of references also mattered. While one expert mentioned “for last sentence could have used any 1 of the 3 refs given, why all 3?” another wrote “stylistically just not done to have paragraphs 1 and 2 be a series of facts from a single reference like this.” Moreover, it was considered problematic when references “are unavailable and/or not in English.”

Further, as medicine is a rapidly growing area, "up to date" information and “current guidelines” were viewed as crucial. It is important to recognize that previously accepted practices may become contraindicated as new evidence emerges. While human experts are typically aware of such shifts, LLMs do not consistently verify whether previously retrieved information has been updated. Importantly, summaries should include details "drawn from the body of the text" and should not "borderline-plagiarize from the abstracts."

\subsubsection{Other factors}
For controversial topics, expert opinions and judgment that sometimes "did not have a reference" also mattered. Those summaries went "above and beyond in explaining more than basic contraindications," included "clarifications" such as "it is too premature to draw any definitive conclusion," and mentioned "ongoing trials." Furthermore, it is important not to carry forward "wrong information," "if the source was wrong," and "doses of drugs that are not clinically relevant" (e.g., doses that were studied in clinical trials but not approved for clinical use).

The comments on literature highlighted different types of errors expected in human writing and LLM outputs. Human errors included "extensive grammatical errors", and "a missing word" (e.g., "inhalation of 100\%", where it should say "inhalation of 100\% O2"), and experts pointed out "poor readability / choppy flow". On the other hand, reviewers mentioned “lack of mistakes” and “absence of errors and omissions” as features of AI-suspected summaries. They identified several types of errors, including instances in which the model incorrectly referred to its output as a review or systematic review.

\section{Limitations}
This study has several notable strengths, including extensive, detailed, and critical evaluations by headache specialists using standardized quantitative rubrics complemented by qualitative free-text feedback, as well as a comprehensive comparison between high-quality expert-written summaries and outputs from three distinct LLM models. There are several limitations, including the relatively small sample size of 10 questions and 10 headache specialist evaluators, which might limit the ability to generalize the findings to other specialties or to literature intended for non-professional audiences. However, due to the considerable time and effort required for detailed annotation, coupled with the active clinical and research commitments of the participating headache specialists, increasing the sample size was not feasible. Furthermore, as the evaluated summaries were generated in early 2025, subsequent advancements in LLMs may have led to improved performance.

\section{Conclusion}
We conducted a comprehensive study comparing literature summaries generated by RAG-based LLMs within a pre-developed agentic chain-of-LLMs pipeline to human expert-written summaries, with critical evaluation by headache specialists. While standard quantitative metrics reveal only a small gap between human experts and LLMs, free-form expert evaluations uncover critical qualitative differences, particularly in the clinical relevance and interpretive depth of the summaries, that LLMs often fail to capture. This highlights a substantial and previously uncharacterized gap between model-generated and human-authored outputs. Our findings further identify expert-valued key features, beyond conventional evaluation metrics, that can inform the future refinement of both human and AI-driven literature summarization workflows.




\subsection*{Conflict of Interest}
JHr owns stock in Amgen and Stryker, serves on Advisory Board for Pfizer, and receives funding from Department of Health and Human Services (K23NS130143). PZ serves on Advisory Board for Pfizer and as a consulttant for Acumen LLC. KI receives Postdoctoral Fellowship Grant from the American Heart Association with funds paid to her institution (https://doi.org/10.58275/AHA.25POST1378529.pc.gr.227396). CC has served as a consultant for: Pfizer, AbbVie, Amneal, Satsuma, and eNeura. She receives research funds from the American Heart Association, Pfizer, Lundbeck and the National Institute of Health with funds paid to her institution. Within the prior 48 months, TJS has served as a consultant with AbbVie, Lundbeck, and Salvia. He received royalties from UpToDate and has stock options/equity interest in Allevalux and Nocira. His institution received research grant funding on his behalf from AbbVie, American Heart Association, Flinn Foundation, Henry Jackson Foundation, National Headache Foundation, National Institutes of Health, Patient Centered Outcomes Research Institute, Pfizer, and the United States Department of Defense. GD and FMC report no conflicting interests. Jennifer Stern is a shareholder in Emmyon Inc, which has no relevance to any topic in this study

\acks{
This work was supported in part by the Clinical Excellence Research Center at Stanford Medicine and Technology and Digital Solutions at Stanford Healthcare. MW is supported by an HAI Graduate Fellowship. AL is supported by an ARC Institute Graduate Fellowship. KI is supported by the American Heart Association Postdoctoral Fellowship, with the funds paid to Mayo Clinic.
}

\bibliography{bib.bib}

\newpage
\appendix
\newpage

\begin{tcolorbox}[
    colback=blue!3,     
    colframe=blue!60!black,
    title=\textbf{Email for participation},
    fonttitle=\bfseries,
    sharp corners,
    boxrule=0.5pt,
    breakable
]

I'm writing to see if you would be interested in participating in a new AI research study I'm conducting. 

\medskip

This study is about using Large Language Models (LLMs) to summarize literature published in PubMed and will involve 10 headache specialists.

\medskip

In the first phase, each headache specialist will be invited to write a clinical question and an answer to that question. The answer will be around 200--300 words, in a review article format with citations — imagine you are writing a brief review for \textit{UpToDate} or for a reputed journal. The readers would be clinicians and providers, and the information is to be used to guide their clinical practice.

\medskip

In the second phase, you will be asked to evaluate the responses/answers written by 9 other headache specialists as well as AI-generated summaries from 4 different LLM systems. You will be blinded to whether the response was written by a human expert or AI, and we will ask you to evaluate whether each summary is correct and useful, and your preference/reasoning of preference for those summaries.

\medskip

If you are interested in participating, we will send out additional details and instructions. If for any reason you are not able to, I'd totally understand! Please keep this confidential.

\medskip
Thank you for considering this request,
\end{tcolorbox}

\begin{tcolorbox}[
    colback=blue!3,     
    colframe=blue!60!black,
    title=\textbf{Detailed instruction on writing the paragraph},
    fonttitle=\bfseries,
    sharp corners,
    boxrule=0.5pt,
    breakable
]

Thank you again for agreeing to participate in the Large Language Model (LLM) Literature Summary for Headache Medicine Project. The goal of this project is to evaluate AI-generated summaries vs human expert-written summaries. Through this project, we want to identify the key parameters that human experts and readers value when reading literature summaries, which would benefit future efforts using AI for summarization and evaluation of LLM-generated outputs.

\medskip

As a brief outline of this project, I aim to involve 10 headache specialists.
In the first phase, each headache specialist will be invited to write one clinical question and an answer to that question. The answer (summary) should be 200-300 words, in a review format with around 10 citations/ references- imagine you are writing a brief review for UpToDate or a reputed journal, and what you write should be useful for clinicians, providers, trainees, and medical students to guide their clinical practice and medical education. As a reminder, your written summary, along with AI-generated summaries, will be judged by other colleagues for correctness, usefulness, and preference, so please refrain from writing opinions that might be too controversial. Also, please do NOT use ChatGPT or any other commercially available AI literature summarization website during the writing process. Using Grammarly for grammar check, if needed, is allowed, but please do not put your ideas, sentences, or written summary into ChatGPT or other LLMs.

\medskip

In the second phase, you will be asked to evaluate the responses/answers written by 9 other headache specialists and AI-generated summaries from different LLM systems. You will be blinded to whether the response was written by a human or AI, and we will ask you to evaluate whether each summary is correct and useful, and your preference for those summaries, and the reason why.
For Step 1, I’d appreciate if you could write an answer to the following question:
- “What are the treatment options for cluster headache?" 
Please feel free to rephrase the question or change it to another topic that might be more interesting to you. Please just let me know ASAP if you'd like to change the proposed question, as we would need to use the same wording of these questions for different AI systems to generate a summary.

\medskip

Please let me know if the question wording is acceptable. I'd appreciate if you could send me the 200-300 word summary with citations before [date].

\medskip

Thank you very much again for your help and please let me know if you have any questions!
If for any reason you are no longer able to participate, I'd totally understand. Please let me know if that's the case.

\end{tcolorbox}

\begin{tcolorbox}[
    colback=blue!3,     
    colframe=blue!60!black,
    title=\textbf{Instruction for Evaluation},
    fonttitle=\bfseries,
    sharp corners,
    boxrule=0.5pt,
    breakable
]

Thank you so much for agreeing to participate in this project and providing an excellent summary of the question assigned to you!
 
After multiple rounds of optimization, we have produced outputs from different LLM systems. Now, we are ready to move forward with the second phase- evaluation.

This folder includes all 10 questions and summaries. You will see four summaries for each question. One summary was written by a human headache specialist, and three were generated by different AI/LLM systems. They were shuffled to appear in random order.
 
[link]
 
We would appreciate your help in evaluating the summaries of the following 5 questions assigned to you, highlighted in Yellow below (You only need to fill out the Google Evaluation Form below for these 5 questions and the corresponding summaries)

The evaluation form is here [link to evaluation form]

Please read through the detailed instructions regarding the evaluation as outlined in the Google Form.

Please select the question first, and fill out your evaluations accordingly.
 
I'd suggest opening one window (screen) for the PDF summary, and another window (screen) to fill out the evaluation form. Or, if you'd prefer, print out the PDF summaries for the 5 questions assigned to you. The estimated time to complete this evaluation for 5 questions is approximately 5 hours. 

We would be very grateful if you could complete this before [date]. Please let me know if you have any questions or concerns. Thank you again for your contribution to this project!

\end{tcolorbox}

\begin{table}[htbp]
\centering
\scriptsize
\begin{tabular}{|p{2.8cm}|p{6.2cm}|p{6.2cm}|}
\hline
\textbf{Topics} & \textbf{Correct Responses} & \textbf{Incorrect Responses} \\ \hline

\textbf{Correctness} &
- The content was the most correct. 
- There is some failure to provide citations, but overall, the information provided is correct. 
- Summary D met all the metrics. &
\\ \hline

\textbf{Completeness} &
- Summary is well rounded.

- It seems to be the most comprehensive, considering all aspects of diagnosing and treating cervicogenic headache. 
- It was also the most complete (other summaries left out several contraindications). 
- Discussed pregnancy in detail; mentions all the gepants and mAbs. 
- Reasonably complete. 
- AI did very well on two of them. 
- Summary D met all the metrics. &
- It was somewhat close but b and c were similar. But b was more complete than c. Also d lacked significant content and was out of date up to 2019. 
- d also included pediatric information. 
- Comprehensive of all the reviews. 
- Has lots of details not shown in others. \\ \hline

\textbf{Conciseness} &
- No redundant material. 
- Choice B is close, but it is very long and redundant. 
- Concise and beautifully written with things I didn't know about. 
- Of good length and the one I liked the most. 
- This summary does not have a lot of repetitive padding. &
- It is concise and well written. 
- It was a hard choice between B and C. B was chosen for conciseness. 
- d I think, but a is close second; summary d was best in content, though a little long. 
- I think it is the most concise. 
- Long. \\ \hline

\textbf{Flow} &
- Nice flow that mimicked a natural reading / learning style. 
- Best flow of ideas and medically accurate paragraph on medical therapies. 
- Grouped by medical then surgical management. 
- Focused on helpful aspects for clinicians. 
- Logical introduction and conclusion. &
- Summary a follows a more logical flow, similar to how a clinician thinks. 
- Best job introducing the topic, specifics, and conclusion. 
- There is a beginning, middle, and end. Logical progression. \\ \hline

\textbf{Focus} &
- Useful dosing information. 
- Included key clinical concepts. 
- Differentiates CGRP monoclonal antibodies vs small molecule antagonists. 
- Discusses rare side effects. 
- Addresses controversies relevant to clinical practice. 
- Summary d focuses more on diagnosis. &
- Choice of data from specific studies instead of summarizing. 
- Includes multiple aspects relevant to clinical practice. 
- Good balance between citing data and clinical implications. 
- Answers the prompt better than others. \\ \hline

\textbf{Expert Opinion / Judgement} &
- Included medications though one lacked a reference. 
- References missing for statements that sound like expert opinion. 
- Provides extra context such as serotonin syndrome explanation. 
- Goes above and beyond in explaining contraindications. 
- Includes clinical tips like prioritizing life-threatening disorders. 
- Mentions ongoing trials and updated guidelines. &
- Emphasized identifying emergent causes; felt more clinical judgment. \\ \hline

\textbf{AI Mistakes} &
- Lack of mistakes. 
- Absence of errors and omissions. 
- Avoids phrases like “this systematic review includes.” 
- Does not incorrectly refer to itself as a review. &
\\ \hline

\textbf{Human Errors} &
- Extensive grammatical errors. 
- A few capitalization and grammatical choices that seem human. 
- Missing word example: “inhalation of 100\% O2.” &
- “A” was well-written but missed part of the question—more typical of human error. \\ \hline

\textbf{Reference Quality} &
- High quality references cited. 
- Cites guideline and practice parameter. 
- Multiple references per fact. 
- Used authoritative and EBM-level sources. &
- Did not mention systematic review. 
- Nice use of references. 
- “a” lacked key reference used by others. 
- References were great. \\ \hline

\textbf{Synthesization} &
- Interprets data as a clinician rather than quoting it. 
- Requires inference to summarize citations, showing conceptual understanding. 
- Writing follows a unified conceptual framework. &
\\ \hline

\textbf{Up to Date / Full-text Content} &
- Written like a clinician and up to date. 
- More current than others. 
- Draws ideas from body of text, not just abstracts. &
\\ \hline

\textbf{Readability / Writing Style} &
- Readability was very natural. 
- Reflexive language suggests a human writer. 
- Writing style feels personal. &
- Simple to follow writing style. \\ \hline

\textbf{No Comment} &
- d &
- Had difficulty identifying human summary, but this was best estimate.
- Summary c. \\ \hline

\end{tabular}
\caption{Qualitative feedback across evaluation topics comparing correct and incorrect responses.}
\label{tab:qualitative_feedback}
\end{table}


\begin{table}[htbp]
\centering
\scriptsize
\begin{tabular}{|p{3.3cm}|p{12cm}|}
\hline
\textbf{Topics} & \textbf{Comments} \\ \hline

\textbf{Synthesization} &
1) Just listing stuff rather than making recommendations. 

3) See my write. 

- AI generated summaries seemed to have “cut and pasted” sentences from the references as opposed to a formulation of the information. 
- Makes sense; metrics only get you so far. \\ \hline

\textbf{Availability of English References} &
- Using references not in English and/or not easily available.  
- Cannot assess use of first 3 references completely as they are unavailable or not in English.  
- All references were legitimate, even reference 1 in Case B which needed the DOI to find it online in PubMed.  
- Summary C uses a non-English-language-only paper where there are good alternatives — possibly indicative of AI.  
- Summary A has a foreign language paper.  
- Metrics helped her but some reference PDFs were missing; had to go to PubMed to find them. \\ \hline

\textbf{Reference Quality} &
- Using weird/old references (e.g., ref 10 is an odd choice focused on three-peat GKS).  
- B and D listed systematic reviews.  
- The current metrics don't capture reference quality, but it was a distinguishing characteristic.  
- Good references, including a major meta-analysis and guidelines for specialists and non-specialists.  
- Excellent authorship and inclusion of imaging references. \\ \hline

\textbf{Focus} &
- Balanced; one sample was oddly surgery-heavy.  
- Passage D full of clinically relevant tips, such as specific triptan considerations and controversies (e.g., SSRIs, hemiplegic migraine).  
- AI-generated ones skew toward diagnosis (strong diagnostic bias).  
- Best summary for broad audience—clinicians, trainees, students—covering diagnosis and management.  
- Metrics not necessarily helpful, but clinically it spoke to how to make diagnosis and treatment decisions.  
- Selection of information not guided by human judgment; summary B seemed bland and non-specific.  
- I was looking for an approach, not therapy; otherwise summary B might have been best. \\ \hline

\textbf{Conciseness} &
- Not repetitive; directness.  
- Introduction and conclusion paragraphs often seem formulaic, redundant, and unhelpful.  
- Concise sentences; great cases so far, but unclear if metrics influenced length.  
- It is concise and well written. \\ \hline

\textbf{Up to Date Information} &
- Citing current guidelines. \\ \hline

\textbf{Number of References} &
- For the second to last sentence could have used ref 5 alone; for the last, any 1 of 3 refs would suffice.  
- Not sure why multiple redundant references used (e.g., MVD as gold, refs 5–8).  
- Unsure why several citations are clustered for single statements.  
- Should use fewer references overall.  
- Summary C uses only 3 references—odd stylistically for a human—but includes Cheema et al.; Farb paper citation unusual. \\ \hline

\textbf{Flow} &
- Layout.  
- Flow of information and logical progression needed.  
- Metrics not that helpful; organization could improve. \\ \hline

\textbf{Human Writing Style} &
- Summary C had the best quality information but poor readability and choppy flow.  
- The writing style of D feels human. \\ \hline

\textbf{Information Noise: Incorrect or Clinically Irrelevant Sources} &
- AI summaries quoted sources accurately but sometimes carried forward incorrect information from poor sources; limited source discernment.  
- Summaries B and C included clinically irrelevant drug doses (e.g., atogepant 120 mg, eptinezumab 30 mg). \\ \hline

\textbf{AI Cliché} &
- A and D read like AI-generated responses.  
- Example: phrases like “This systematic review aims to summarize the evidence-based options…” are typical of LLMs.  
- Two responses use language such as “this review” or “this systematic review,” which was not requested.  
- Noticeable clichés in AI-generated answers. \\ \hline

\textbf{Credit / Authority of Reference Source} &
- Pattern recognition; can be done quickly but one must verify data.  
- By and large good references, but the human one chosen had the most authoritative headache experts and top journals.  
- Ref 6 not very authoritative—likely AI. \\ \hline

\textbf{Expert Opinion / Judgement} &
- Answer A very convincing and detailed; seemed human until reading D, which included reflection and judgment. \\ \hline

\textbf{Others} &
- Whether it was useful and appealing mattered more.  
- Metrics are time-consuming and not always helpful.  
- Allowed a broad interpretation of “evidence-based” to include narrative reviews and clinical experience, aligning with modern EBM definitions.  
- Broader input of assessments may be more valuable. \\ \hline

\end{tabular}
\caption{Expert comments grouped by qualitative topic for AI vs human summary evaluation.}
\label{tab:expert_comments}
\end{table}


\begin{table}[htbp]
\centering
\caption{Correctness: ANOVA and t-Test Results}
\begin{tabular}{lcccccc}
\hline
\textbf{SUMMARY} & \textbf{Count} & \textbf{Sum} & \textbf{Average} & \textbf{Variance} & & \\ \hline
Human & 50 & 457 & 9.14 & 2.61 & & \\
Sonnet & 50 & 456 & 9.12 & 1.82 & & \\
GPT-4o & 50 & 408 & 8.16 & 4.83 & & \\
Llama 3.1 & 50 & 428 & 8.56 & 2.01 & & \\ \hline
\textbf{ANOVA} & \textbf{SS} & \textbf{df} & \textbf{MS} & \textbf{F} & \textbf{P-value} & \textbf{F crit} \\ \hline
Between Groups & 33.655 & 3 & 11.218 & 3.981 & 0.009 & 2.651 \\
Within Groups & 552.340 & 196 & 2.818 &  &  &  \\
Total & 585.995 & 199 &  &  &  &  \\ \hline
\end{tabular}

\vspace{0.5em}
\begin{tabular}{lcccccc}
\multicolumn{7}{c}{\textbf{t-Test: Human vs. Sonnet}} \\ \hline
 & \textbf{Human} & \textbf{Sonnet} &  &  &  &  \\
Mean & 9.14 & 9.12 &  &  &  &  \\
Variance & 2.61 & 1.82 &  &  &  &  \\
Observations & 50 & 50 &  &  &  &  \\
Pooled Variance & 2.22 &  &  &  &  &  \\
df & 98 &  &  &  &  &  \\
t Stat & 0.07 &  &  &  &  &  \\
P(T$\le$t) one-tail & 0.47 &  &  &  &  &  \\
t Critical one-tail & 1.66 &  &  &  &  &  \\
P(T$\le$t) two-tail & 0.95 &  &  &  &  &  \\
t Critical two-tail & 1.98 &  &  &  &  &  \\ \hline
\multicolumn{7}{c}{\textbf{t-Test: Human vs. GPT-4o}} \\ \hline
Mean & 9.14 & 8.16 &  &  &  &  \\
Variance & 2.61 & 4.83 &  &  &  &  \\
Observations & 50 & 50 &  &  &  &  \\
Pooled Variance & 3.72 &  &  &  &  &  \\
df & 98 &  &  &  &  &  \\
t Stat & 2.54 &  &  &  &  &  \\
P(T$\le$t) one-tail & 0.01 &  &  &  &  &  \\
t Critical one-tail & 1.66 &  &  &  &  &  \\
P(T$\le$t) two-tail & 0.01 &  &  &  &  &  \\
t Critical two-tail & 1.98 &  &  &  &  &  \\ \hline
\multicolumn{7}{c}{\textbf{t-Test: Human vs. Llama 3.1}} \\ \hline
Mean & 9.14 & 8.56 &  &  &  &  \\
Variance & 2.61 & 2.01 &  &  &  &  \\
Observations & 50 & 50 &  &  &  &  \\
Pooled Variance & 2.31 &  &  &  &  &  \\
df & 98 &  &  &  &  &  \\
t Stat & 1.91 &  &  &  &  &  \\
P(T$\le$t) one-tail & 0.03 &  &  &  &  &  \\
t Critical one-tail & 1.66 &  &  &  &  &  \\
P(T$\le$t) two-tail & 0.06 &  &  &  &  &  \\
t Critical two-tail & 1.98 &  &  &  &  &  \\ \hline
\end{tabular}
\end{table}


\begin{table}[htbp]
\centering
\caption{Completeness: ANOVA and t-Test Results}
\begin{tabular}{lcccccc}
\hline
\textbf{SUMMARY} & \textbf{Count} & \textbf{Sum} & \textbf{Average} & \textbf{Variance} & & \\ \hline
Human & 50 & 443 & 8.86 & 1.35 & & \\
Sonnet & 50 & 420 & 8.40 & 3.18 & & \\
GPT-4o & 50 & 413 & 8.26 & 2.11 & & \\
Llama 3.1 & 50 & 393 & 7.86 & 2.82 & & \\ \hline
\textbf{ANOVA} & \textbf{SS} & \textbf{df} & \textbf{MS} & \textbf{F} & \textbf{P-value} & \textbf{F crit} \\ \hline
Between Groups & 25.54 & 3 & 8.51 & 3.60 & 0.015 & 2.65 \\
Within Groups & 463.66 & 196 & 2.37 &  &  &  \\
Total & 489.20 & 199 &  &  &  &  \\ \hline
\end{tabular}

\vspace{0.5em}
\begin{tabular}{lcccccc}
\multicolumn{7}{c}{\textbf{t-Test: Human vs. Sonnet}} \\ \hline
 & \textbf{Human} & \textbf{Sonnet} &  &  &  &  \\
Mean & 8.86 & 8.40 &  &  &  &  \\
Variance & 1.35 & 3.18 &  &  &  &  \\
Observations & 50 & 50 &  &  &  &  \\
Pooled Variance & 2.27 &  &  &  &  &  \\
df & 98 &  &  &  &  &  \\
t Stat & 1.53 &  &  &  &  &  \\
P(T$\le$t) one-tail & 0.06 &  &  &  &  &  \\
t Critical one-tail & 1.66 &  &  &  &  &  \\
P(T$\le$t) two-tail & 0.13 &  &  &  &  &  \\
t Critical two-tail & 1.98 &  &  &  &  &  \\ \hline
\multicolumn{7}{c}{\textbf{t-Test: Human vs. GPT-4o}} \\ \hline
Mean & 8.86 & 8.26 &  &  &  &  \\
Variance & 1.35 & 2.11 &  &  &  &  \\
Observations & 50 & 50 &  &  &  &  \\
Pooled Variance & 1.73 &  &  &  &  &  \\
df & 98 &  &  &  &  &  \\
t Stat & 2.28 &  &  &  &  &  \\
P(T$\le$t) one-tail & 0.01 &  &  &  &  &  \\
t Critical one-tail & 1.66 &  &  &  &  &  \\
P(T$\le$t) two-tail & 0.02 &  &  &  &  &  \\
t Critical two-tail & 1.98 &  &  &  &  &  \\ \hline
\multicolumn{7}{c}{\textbf{t-Test: Human vs. Llama 3.1}} \\ \hline
Mean & 8.86 & 7.86 &  &  &  &  \\
Variance & 1.35 & 2.82 &  &  &  &  \\
Observations & 50 & 50 &  &  &  &  \\
Pooled Variance & 2.08 &  &  &  &  &  \\
df & 98 &  &  &  &  &  \\
t Stat & 3.47 &  &  &  &  &  \\
P(T$\le$t) one-tail & 0.00 &  &  &  &  &  \\
t Critical one-tail & 1.66 &  &  &  &  &  \\
P(T$\le$t) two-tail & 0.00 &  &  &  &  &  \\
t Critical two-tail & 1.98 &  &  &  &  &  \\ \hline
\end{tabular}
\end{table}



\begin{table}[htbp]
\centering
\caption{Conciseness: ANOVA and t-Test Results}
\begin{tabular}{lcccccc}
\hline
\textbf{SUMMARY} & \textbf{Count} & \textbf{Sum} & \textbf{Average} & \textbf{Variance} & & \\ \hline
Human & 50 & 476 & 9.52 & 2.30 & & \\
Sonnet & 50 & 476 & 9.52 & 0.62 & & \\
GPT-4o & 50 & 438 & 8.76 & 1.70 & & \\
Llama 3.1 & 50 & 424 & 8.48 & 1.81 & & \\ \hline
\textbf{ANOVA} & \textbf{SS} & \textbf{df} & \textbf{MS} & \textbf{F} & \textbf{P-value} & \textbf{F crit} \\ \hline
Between Groups & 42.46 & 3 & 14.15 & 8.82 & 0.000016 & 2.65 \\
Within Groups & 314.56 & 196 & 1.60 &  &  &  \\
Total & 357.02 & 199 &  &  &  &  \\ \hline
\end{tabular}

\vspace{0.5em}
\begin{tabular}{lcccccc}
\multicolumn{7}{c}{\textbf{t-Test: Human vs. Sonnet}} \\ \hline
 & \textbf{Human} & \textbf{Sonnet} &  &  &  &  \\
Mean & 9.52 & 9.52 &  &  &  &  \\
Variance & 2.30 & 0.62 &  &  &  &  \\
Observations & 50 & 50 &  &  &  &  \\
Pooled Variance & 1.46 &  &  &  &  &  \\
df & 98 &  &  &  &  &  \\
t Stat & 0.00 &  &  &  &  &  \\
P(T$\le$t) one-tail & 0.50 &  &  &  &  &  \\
t Critical one-tail & 1.66 &  &  &  &  &  \\
P(T$\le$t) two-tail & 1.00 &  &  &  &  &  \\
t Critical two-tail & 1.98 &  &  &  &  &  \\ \hline
\multicolumn{7}{c}{\textbf{t-Test: Human vs. GPT-4o}} \\ \hline
Mean & 9.52 & 8.76 &  &  &  &  \\
Variance & 2.30 & 1.70 &  &  &  &  \\
Observations & 50 & 50 &  &  &  &  \\
Pooled Variance & 2.00 &  &  &  &  &  \\
df & 98 &  &  &  &  &  \\
t Stat & 2.69 &  &  &  &  &  \\
P(T$\le$t) one-tail & 0.00 &  &  &  &  &  \\
t Critical one-tail & 1.66 &  &  &  &  &  \\
P(T$\le$t) two-tail & 0.01 &  &  &  &  &  \\
t Critical two-tail & 1.98 &  &  &  &  &  \\ \hline
\multicolumn{7}{c}{\textbf{t-Test: Human vs. Llama 3.1}} \\ \hline
Mean & 9.52 & 8.48 &  &  &  &  \\
Variance & 2.30 & 1.81 &  &  &  &  \\
Observations & 50 & 50 &  &  &  &  \\
Pooled Variance & 2.05 &  &  &  &  &  \\
df & 98 &  &  &  &  &  \\
t Stat & 3.63 &  &  &  &  &  \\
P(T$\le$t) one-tail & 0.00 &  &  &  &  &  \\
t Critical one-tail & 1.66 &  &  &  &  &  \\
P(T$\le$t) two-tail & 0.00 &  &  &  &  &  \\
t Critical two-tail & 1.98 &  &  &  &  &  \\ \hline
\end{tabular}
\end{table}



\begin{table}[htbp]
\centering
\caption{Usefulness: ANOVA and t-Test Results}
\begin{tabular}{lcccccc}
\hline
\textbf{SUMMARY} & \textbf{Count} & \textbf{Sum} & \textbf{Average} & \textbf{Variance} & & \\ \hline
Human & 50 & 407 & 8.14 & 3.96 & & \\
Sonnet & 50 & 367 & 7.34 & 4.64 & & \\
GPT-4o & 50 & 330 & 6.60 & 5.18 & & \\
Llama 3.1 & 50 & 306 & 6.12 & 5.01 & & \\ \hline
\textbf{ANOVA} & \textbf{SS} & \textbf{df} & \textbf{MS} & \textbf{F} & \textbf{P-value} & \textbf{F crit} \\ \hline
Between Groups & 116.98 & 3 & 38.99 & 8.30 & 3.17E-05 & 2.65 \\
Within Groups & 920.52 & 196 & 4.70 &  &  &  \\
Total & 1037.50 & 199 &  &  &  &  \\ \hline
\end{tabular}

\vspace{0.5em}
\begin{tabular}{lcccccc}
\multicolumn{7}{c}{\textbf{t-Test: Human vs. Sonnet}} \\ \hline
 & \textbf{Human} & \textbf{Sonnet} &  &  &  &  \\
Mean & 8.14 & 7.34 &  &  &  &  \\
Variance & 3.96 & 4.64 &  &  &  &  \\
Observations & 50 & 50 &  &  &  &  \\
Pooled Variance & 4.30 &  &  &  &  &  \\
df & 98 &  &  &  &  &  \\
t Stat & 1.93 &  &  &  &  &  \\
P(T$\le$t) one-tail & 0.03 &  &  &  &  &  \\
t Critical one-tail & 1.66 &  &  &  &  &  \\
P(T$\le$t) two-tail & 0.06 &  &  &  &  &  \\
t Critical two-tail & 1.98 &  &  &  &  &  \\ \hline
\multicolumn{7}{c}{\textbf{t-Test: Human vs. GPT-4o}} \\ \hline
Mean & 8.14 & 6.60 &  &  &  &  \\
Variance & 3.96 & 5.18 &  &  &  &  \\
Observations & 50 & 50 &  &  &  &  \\
Pooled Variance & 4.57 &  &  &  &  &  \\
df & 98 &  &  &  &  &  \\
t Stat & 3.60 &  &  &  &  &  \\
P(T$\le$t) one-tail & 0.00 &  &  &  &  &  \\
t Critical one-tail & 1.66 &  &  &  &  &  \\
P(T$\le$t) two-tail & 0.00 &  &  &  &  &  \\
t Critical two-tail & 1.98 &  &  &  &  &  \\ \hline
\multicolumn{7}{c}{\textbf{t-Test: Human vs. Llama 3.1}} \\ \hline
Mean & 8.14 & 6.12 &  &  &  &  \\
Variance & 3.96 & 5.01 &  &  &  &  \\
Observations & 50 & 50 &  &  &  &  \\
Pooled Variance & 4.48 &  &  &  &  &  \\
df & 98 &  &  &  &  &  \\
t Stat & 4.77 &  &  &  &  &  \\
P(T$\le$t) one-tail & 0.00 &  &  &  &  &  \\
t Critical one-tail & 1.66 &  &  &  &  &  \\
P(T$\le$t) two-tail & 0.00 &  &  &  &  &  \\
t Critical two-tail & 1.98 &  &  &  &  &  \\ \hline
\end{tabular}
\end{table}

\end{document}